%% file: paper.tex
\documentclass[]{bytedance_seed}

\usepackage[toc,page,header]{appendix}

\usepackage{minitoc}
\usepackage{amsmath}
\usepackage{amsfonts}

\usepackage{tcolorbox}
\usepackage{courier}   
\usepackage{xcolor}

\tcbset{
  promptbox/.style={
    colback=blue!3,
    colframe=blue!60,
    coltitle=white,
    fonttitle=\bfseries,
    boxrule=0.8pt,
    arc=4pt,
    left=6pt,
    right=6pt,
    top=6pt,
    bottom=6pt,
  }
}

\tcbset{
  problembox/.style={
    colback=green!5,
    colframe=green!60!black,
    coltitle=white,
    fonttitle=\bfseries,
    boxrule=0.8pt,
    arc=4pt,
    left=6pt,
    right=6pt,
    top=6pt,
    bottom=6pt,
  }
}

\title{Learn Hard Problems During RL with Reference Guided Fine-tuning}

\author[2,*,\dagger]{Yangzhen Wu}
\author[3]{Shanda Li}
\author[3]{Zixin Wen}
\author[1]{Xin zhou}
\author[3]{Ameet Talwalkar}
\author[3]{Yiming Yang}
\author[1,\dagger]{Wenhao Huang}
\author[1,\dagger]{Tianle Cai}

\affiliation[1]{ByteDance Seed}
\affiliation[2]{UC Berkeley}
\affiliation[3]{Carnegie Mellon University}

\contribution[*]{Work done during internship at ByteDance Seed}
\contribution[\dagger]{Corresponding authors}

\abstract{
Reinforcement learning (RL) for mathematical reasoning can suffer from reward sparsity: for challenging problems, LLM fails to sample any correct trajectories, preventing RL from receiving meaningful positive feedback. At the same time, there often exists human written reference solutions along with the problem (e.g. problems from AoPs), but directly fine-tuning on these solutions offers no benefit because models often cannot imitate human proofs that lie outside their own reasoning distribution. We introduce reference-guided fine-tuning (ReGFT), a simple and effective method that utilizes human-written reference solutions to synthesize positive trajectories on hard problem and train on them before RL. For each problem, we provide the model with partial reference solution and let it to generate its own reasoning trace, ensuring the resulting trajectories remain in the model’s reasoning space while still benefiting from reference guidance. Fine-tuning on these reference guided trajectories increases the number of solvable problems and produces a checkpoint that will get more positive rewards during RL. Across three benchmarks (AIME’24, AIME’25, Beyond-AIME), ReGFT consistently improves supervised accuracy, accelerates DAPO training, and raises the final performance plateau of RL. Our results show that ReGFT can effectively overcome reward sparsity and unlock stronger RL-based mathematical reasoning.
}

\correspondence{Yangzhen Wu at \email{yangzhen\_wu@berkeley.edu}}

\begin{document}
\maketitle

\input{sections/introduction}

\input{sections/relatedwork}
\input{sections/approach}

\input{sections/experiments}

\input{sections/conclusion}

\clearpage

\bibliographystyle{plainnat}
\bibliography{main}

\clearpage

\beginappendix

\input{sections/appendix}

\end{document}

%% file: sections/introduction.tex
\section{Introduction}

Reinforcement learning (RL) has been proven effective for enhancing the reasoning capabilities of large language models (LLMs), as demonstrated by the success of OpenAI’s o1 and DeepSeek’s R1 \citep{openai2024openaio1card, deepseekai2025deepseekr1incentivizingreasoningcapability}. These breakthroughs establish a new paradigm in which increasing computational resources are allocated to post-training, with RL serving as the central mechanism for capability improvement.

Within this paradigm, Reinforcement Learning with Verifiable Rewards (RLVR) has emerged as a central framework for mathematical reasoning. RLVR trains models by sampling multiple reasoning trajectories for each prompt and optimizing the model to increase the likelihood of trajectories that lead to correct solutions. Supervision is provided by an automatic, rule-based verifier that assigns rewards based solely on the correctness of the final answer, eliminating the need for human-label or reward model training. Through this process, RLVR systematically promotes reasoning paths consistent with verifiable correctness while suppressing incorrect ones, thereby eliciting and strengthening the latent reasoning capabilities present in the base model.

Despite its effectiveness, RL remains fundamentally constrained by the competence of the base model. In RLVR, learning signals are obtained only when sampled reasoning trajectories receive positive rewards from the verifier. For problems that exceed the model’s current reasoning capability--such as complex problems or unseen reasoning tasks--the base model often fails to produce any correct trajectories, leading to sparse or zero rewards. In these regimes, training effectively stalls due to the absence of meaningful gradient signals. This limitation renders RL both inefficient and fragile, as substantial computation is wasted on trajectories that provide no informative feedback.

In this work, we investigate how a fixed training dataset can be utilized to maximally elicit a model’s mathematical reasoning capability through reinforcement learning, with a particular focus on mitigating reward sparsity during RL. In a typical training dataset, each problem is accompanied by a reference chain-of-thought (CoT) solution, which is commonly used for supervised fine-tuning (SFT). However, directly fine-tuning on these reference solutions often yields limited performance gains, due in part to a mismatch between the model’s intrinsic reasoning patterns and the structure of the provided reference, leading to poor generalization \citep{luong2024reftreasoningreinforcedfinetuning}. To address this issue, prior work has proposed REinforced Fine-Tuning (ReFT) \citep{luong2024reftreasoningreinforcedfinetuning}, which performs on-policy sampling and applies SFT to model-generated trajectories that are verified as correct. While ReFT can alleviate reward sparsity for problems where the model is already capable of producing correct solutions, it fails to improve learning on harder problems that lie beyond the model’s initial capability, where no correct trajectories can be sampled and reinforcement signals remain absent.

To overcome this limitation, we propose Reference-Guided Finetuning (ReGFT), which explicitly leverages reference solutions to synthesize training trajectories for hard problems that the model cannot solve independently. Instead of directly fine-tuning on reference CoTs, ReGFT uses reference solutions as guidance to elicit model-generated reasoning traces that are aligned with the model’s own reasoning process while remaining solution-consistent. The model is then fine-tuned on a mixture of these reference-guided trajectories and its self-generated reasoning traces. This process substantially increases the model’s pass rate on hard training problems, resulting in denser and more informative reward signals during subsequent reinforcement learning. In experiments, we found that
\begin{itemize}
    \item As shown in Figure~\ref{fig:benchmark_comparison}, models initialized with ReGFT consistently outperform the raw checkpoint throughout RL training across all benchmarks, exhibiting both faster performance gains in early stages and higher final accuracy, indicating that reference-guided fine-tuning effectively alleviates reward sparsity and improves optimization efficiency.
    \item Comparative results show that directly fine-tuning on reference solutions alone is insufficient; instead, learning benefits critically depend on model-derived reasoning trajectories that are guided by references but remain aligned with the model’s own generation process, which is essential for robust learning and generalization.
    \item Analysis of \texttt{pass@k} under increasing inference budgets reveals that RL can improves models' \texttt{pass@k} performance. And ReGFT yields more stable and sustained improvements for all compute budget, while gains from ReFT are primarily limited to low-$k$ regimes and do not scale as effectively.
\end{itemize}

%% file: sections/relatedwork.tex
\section{Related Work}

This work studies a central question in reinforcement learning for language models:
\textit{How can RL enable a model to solve problems that the base model is originally unable to solve?}

Recent studies \citep{yue2025doesreinforcementlearningreally,wu2026invisibleleashrlvrescape} show that naïve RL often fails to expand a model’s intrinsic problem-solving frontier. In particular, they observe that RL training can exhibit \texttt{pass@k} saturation as $k$ increases and may even underperform the model’s raw inference performance on problems that are initially unsolvable. These results suggest that standard RL primarily amplifies existing behaviors rather than inducing genuinely new reasoning capabilities.

Motivated by this limitation, a growing body of work modifies the RL pipeline to enable learning beyond the model’s original competence—specifically, to improve performance on hard or previously unsolvable problems. Existing approaches can be broadly grouped into three complementary directions.

\paragraph{Scaling RL and Adaptive Sampling.}
Several works attribute the limitations of naïve RL to insufficient or poorly allocated exploration rather than a fundamental inability of RL to improve reasoning. These methods improve RL either by prolonging training horizons \citep{liu2025prorlprolongedreinforcementlearning,hu2025brorlscalingreinforcementlearning} or by adaptively reallocating rollout budgets toward harder and more informative instances \citep{li2025knapsackrlunlockingexploration,yang2025depthbreadthsynergyrlvrunlocking}. Concretely, they scale exploration breadth via increased rollouts per example, extend training depth to allow new strategies to emerge, and design difficulty-aware sampling schemes that focus updates on problems with stronger learning signals. Together, these approaches suggest that scaling and redistributing exploration during RL can improve learning on problems where uniform, short-horizon RL shows limited progress.

\paragraph{Question Augmentation during RL.}
Another line of work improves learning on hard problems by modifying the question distribution during RL to provide more informative training signals. \citep{li2025questaexpandingreasoningcapacity} augments difficult questions with partial solutions or scaffolding, increasing the likelihood of sampling rewarding trajectories and improving sample efficiency. \citep{chen2025nudgingboundariesllmreasoning} similarly introduces self-generated or externally provided hints for initially unsolvable questions, transforming zero-reward cases into learnable instances. These approaches demonstrate that augmenting hard questions during RL can help models make progress when standard RL receives little or no learning signal.

\paragraph{Interleaving SFT and RL.}
A third direction combines supervised fine-tuning (SFT) and RL to address the difficulty of discovering correct trajectories on hard problems. \citep{ma2025learningreinforcementlearningcant} explicitly alternates between online SFT and RL, using SFT to introduce solutions on the hardest questions and RL to reinforce and generalize these behaviors. In contrast, \citep{fu2025srftsinglestagemethodsupervised} proposes a single-stage objective that jointly optimizes supervised and reinforcement losses to improve training stability without phase switching. \citep{zhou2026expounlockinghardreasoning} augments RL with self-generated explanations as auxiliary supervision to guide exploration toward structured reasoning trajectories, while \citep{zhang2025breadbranchedrolloutsexpert} anchors RL rollouts to partial expert prefixes when self-generated trajectories fail, ensuring the presence of successful rollouts and densifying the reward signal.

\paragraph{Our Difference.}
Our work differs from the above approaches in both \emph{timing} and \emph{methodology}. While prior methods modify RL itself—by injecting supervised objectives during RL training or guiding exploration with expert-prefixed rollouts—we instead focus on improving the model’s competence \emph{before} RL. Specifically, we investigate whether targeted SFT prior to RL can raise the model’s baseline capability on hard problems such that subsequent RL naturally receives positive rewards on instances that were previously unsolvable. This perspective is supported by observations that stronger mid-training leads to more effective downstream RL \citep{zhang2025interplaypretrainingmidtrainingrl}, and is orthogonal to the choice of RL algorithm.

The closest related works are ReFT \citep{luong2024reftreasoningreinforcedfinetuning} and RL Teachers \citep{cetin2025reinforcementlearningteacherstest}. ReFT fine-tunes models on self-generated correct traces to increase the probability of producing correct trajectories on hard problems, while RL Teachers distill trajectories generated by an RL-trained teacher model. In contrast, our method interleaves the model’s own reasoning with expert solutions available in the training data to synthesize trajectories for hard problems \emph{prior to RL}, thereby improving the model’s initial competence and enabling more effective reward-driven learning in subsequent RL stages.

%% file: sections/approach.tex
\section{Approach}

\begin{figure*}[ht]
    \centering
    \includegraphics[width=0.98\linewidth]{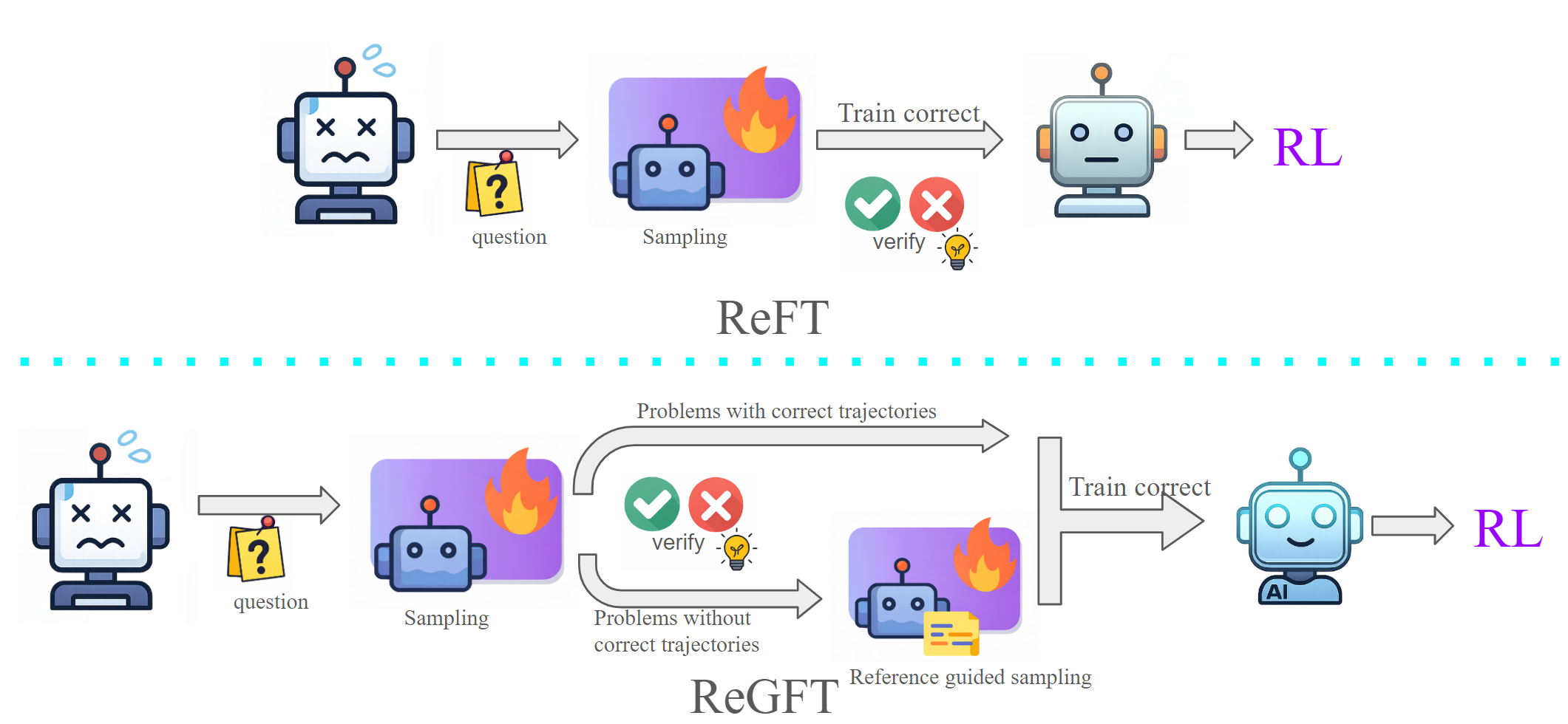}
    \hfill
    \vspace{-2mm}
  \caption{Comparison of ReFT and ReGFT.
\textbf{Top:} ReFT fine-tunes the model using verified correct trajectories obtained from standard sampling.
\textbf{Bottom:} ReGFT additionally applies reference-guided sampling to recover hard problems without correct trajectories.}
    \label{fig:approach}
\end{figure*}

\paragraph{On-policy finetuning with model-generated trajectories (ReFT).}
We first consider REinforced Fine-Tuning (ReFT) \citep{luong2024reftreasoningreinforcedfinetuning}, which applies supervised fine-tuning to correct reasoning trajectories generated by the model itself. By restricting supervision to on-policy, verified trajectories, ReFT aligns training signals with the model’s intrinsic reasoning patterns and avoids the mismatch often observed when directly fine-tuning on human-written reference solutions. This approach is effective for problems where the model is already capable of producing correct trajectories, as it increases the likelihood of such trajectories and improves the model’s readiness for subsequent reinforcement learning. However, ReFT fundamentally depends on the availability of model-generated correct solutions and therefore does not apply to harder problems where the base model fails to produce any correct trajectories.

\paragraph{Reference-Guided Finetuning for Unsolved Problems (ReGFT).}
To extend supervised learning beyond the model’s initial capability boundary, we propose \emph{Reference-Guided Finetuning} (ReGFT). ReGFT is designed to improve the model’s ability to generate correct reasoning trajectories on problems that are initially unsolvable by the base model. Similar to prior work that assumes access to reference solutions in the training data \citep{zhang2025breadbranchedrolloutsexpert,li2025questaexpandingreasoningcapacity,chen2025nudgingboundariesllmreasoning,cetin2025reinforcementlearningteacherstest}, we leverage these references to construct effective and informative training signals for hard problems.

Concretely, ReGFT synthesizes training trajectories by prompting the model with partial reference solutions as hints. These hints provide high-level structural guidance about the solution while requiring the model to reason from the beginning using its own generative process, rather than copying or completing a reference trajectory. This design preserves the model’s intrinsic reasoning style and avoids the brittleness associated with directly fine-tuning on full reference solutions. Unlike BREAD, which injects expert prefixes during RL rollouts, and RL Teachers, which distill trajectories generated by an RL-trained teacher model, ReGFT operates entirely in a pre-RL supervised stage and focuses on improving the model’s initial competence on hard problems. As validated in later sections, maintaining self-generated reasoning is crucial for effective generalization and for enabling downstream reinforcement learning. In practice, we fine-tune the model on a mixture of self-generated correct trajectories (as in ReFT) and reference-guided trajectories constructed with such hints, which increases the model’s pass rate on hard problems and yields a stronger initialization for subsequent RL. An overview of the ReGFT pipeline is shown in Figure~\ref{fig:approach}. We train only on hard problems, defined as those with less than $25\%$ accuracy under the original model when sampled 16 times. Easier problems are excluded to avoid overfitting and to focus learning on cases where the model lacks sufficient competence.

\paragraph{Reinforcement learning from improved initialization}
Starting from checkpoints produced by ReFT or ReGFT, we can apply reinforcement learning to further improve reasoning performance. In practice, ReGFT checkpoints tend to be stronger and are more capable of generating trajectories that receive positive rewards on previously unsolvable questions, making them a better starting point for RL.

Among existing reinforcement learning algorithms for reasoning \citep{schulman2017proximalpolicyoptimizationalgorithms,shao2024deepseekmathpushinglimitsmathematical,yu2025dapo,zheng2025groupsequencepolicyoptimization}, we adopt DAPO (Decoupled Clip and Dynamic sAmpling Policy Optimization) \citep{yu2025dapoopensourcellmreinforcement}, a variant of GRPO\citep{shao2024deepseekmathpushinglimitsmathematical} as our reinforcement learning algorithm. DAPO is a strong and well-validated method that improves training stability and sample efficiency through two key designs: decoupled clipping, which enables flexible reinforcement of high-quality trajectories without collapsing policy diversity, and dynamic sampling, which ensures that each update is based on reward-diverse trajectory groups and thus provides informative learning signals even under sparse rewards.

We deliberately choose DAPO to demonstrate that the benefits of ReGFT are orthogonal to advances in reinforcement learning algorithms. Despite DAPO already incorporating dynamic sampling to mitigate uninformative gradients, initializing training from ReGFT checkpoints consistently leads to further improvements. This shows that ReGFT enables sample more corrected generated trajectories themselves, rather than relying on improved optimization or sampling strategies during RL. Consequently, ReGFT complements even state-of-the-art RL methods like DAPO and can be combined with them to yield additional gains.

%% file: sections/experiments.tex
\section{Experiments}

\subsection{Experimental Setup}

We use Qwen3-4B-2507-Instruct \citep{yang2025qwen3technicalreport} as the base model, which provides strong reasoning performance and reliable instruction following ability. Compared with alternative model families, Qwen3-4B-Instruct offers a more stable foundation in our setting: prior work shows that Qwen2.5 models can be sensitive to spurious reward signals \citep{shao2025spuriousrewardsrethinkingtraining}. We train on OmniMath \citep{gao2024omnimathuniversalolympiadlevel}, which consists of 4,428 Olympiad-level mathematics problems with verified reference solutions and induces severe reward sparsity, making it well suited for studying verifier-based reinforcement learning. For evaluation, we use AIME24\&25 and BeyondAIME \citep{bytedance_seed_2025_beyondaime}, a more challenging and unsaturated benchmark of 100 problems. All experiments are conducted using the verl framework \citep{sheng2024hybridflow}. We set the maximum generation length to 16{,}384 tokens and use temperature $0.7$ with top-$p$ sampling ($p=0.9$) during generation.

\begin{figure*}[t]
    \centering
    \begin{subfigure}{0.32\textwidth}
        \centering
        \includegraphics[width=\linewidth]{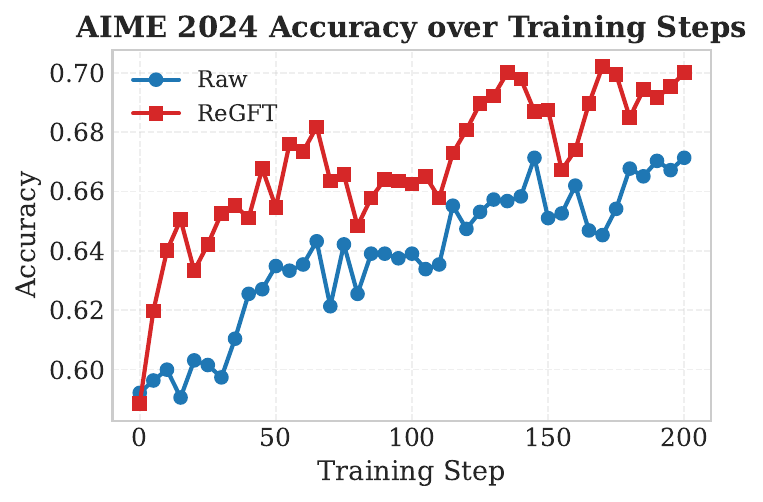}
        \caption{AIME 2024}
        \label{fig:aime24}
    \end{subfigure}
    \hfill
    \begin{subfigure}{0.32\textwidth}
        \centering
        \includegraphics[width=\linewidth]{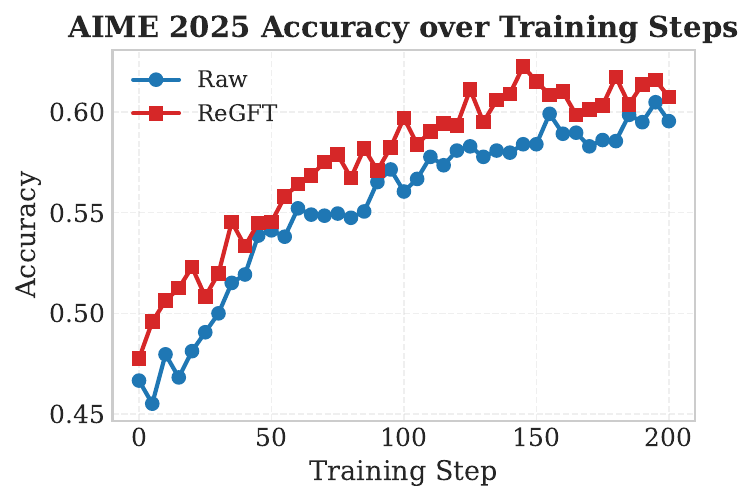}
        \caption{AIME 2025}
        \label{fig:aime25}
    \end{subfigure}
    \hfill
    \begin{subfigure}{0.32\textwidth}
        \centering
        \includegraphics[width=\linewidth]{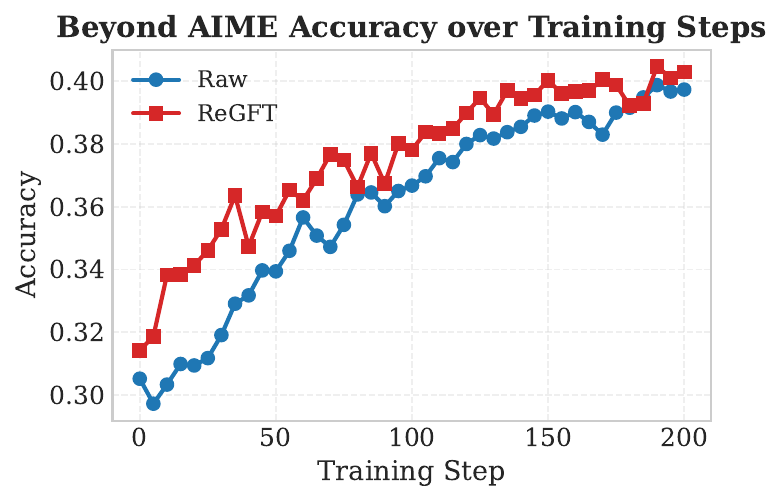}
        \caption{Beyond AIME}
        \label{fig:beyond}
    \end{subfigure}
    
    \vspace{-2mm}
    \caption{
        Reinforcement learning performance over training steps on three challenging benchmarks.
        Models initialized with \textbf{ReGFT} consistently achieve higher accuracy, faster convergence,
        and a superior final performance compared to the raw checkpoint, demonstrating that
        reference-guided fine-tuning provides a stronger initialization for RL.
    }
    \label{fig:benchmark_comparison}
\end{figure*}

\begin{figure*}[t]
    \centering
    \begin{subfigure}{0.32\textwidth}
        \centering
        \includegraphics[width=\linewidth]{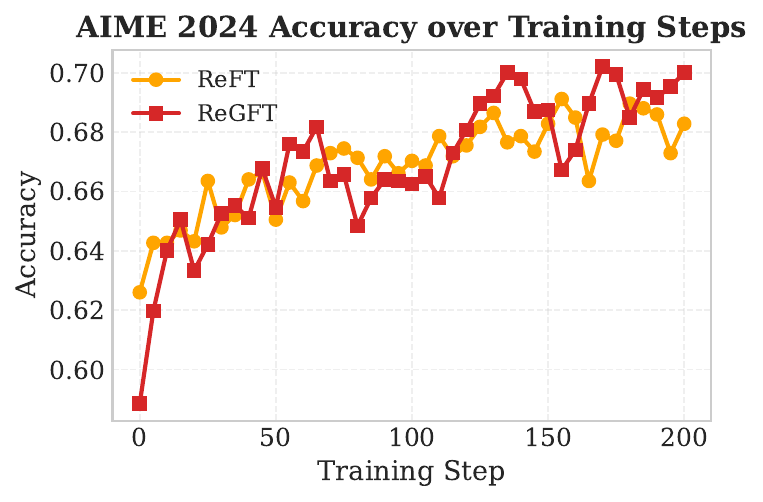}
        \caption{AIME 2024}
        \label{fig:reft_regft_aime24}
    \end{subfigure}
    \hfill
    \begin{subfigure}{0.32\textwidth}
        \centering
        \includegraphics[width=\linewidth]{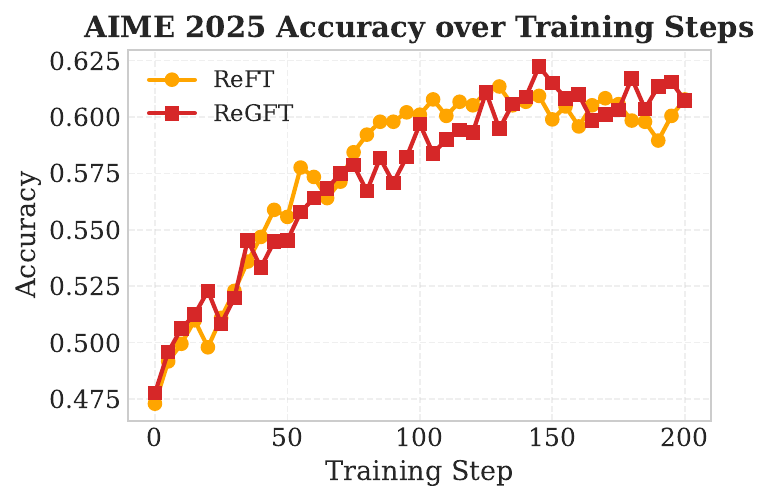}
        \caption{AIME 2025}
        \label{fig:reft_regft_aime25}
    \end{subfigure}
    \hfill
    \begin{subfigure}{0.32\textwidth}
        \centering
        \includegraphics[width=\linewidth]{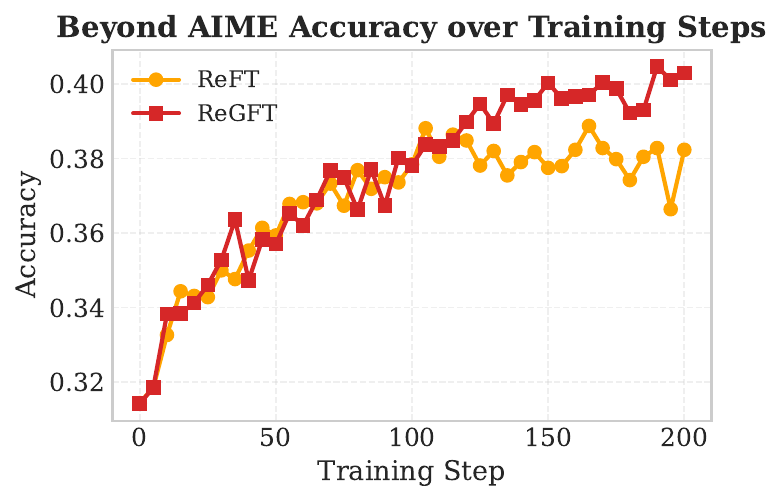}
        \caption{Beyond AIME}
        \label{fig:reft_regft_beyond}
    \end{subfigure}
    
    \vspace{-2mm}
    \caption{
        Reinforcement learning performance comparison between \textbf{ReFT} and \textbf{ReGFT} across three benchmarks.
        While both initializations accelerate early-stage RL, ReGFT consistently achieves higher final accuracy,
        highlighting the contribution of reference-guided demonstrations beyond self-generated trajectories.
    }
    \label{fig:impact_hint_comparison}
\end{figure*}

\begin{figure*}[t]
    \centering
    \begin{subfigure}{0.32\textwidth}
        \centering
        \includegraphics[width=\linewidth]{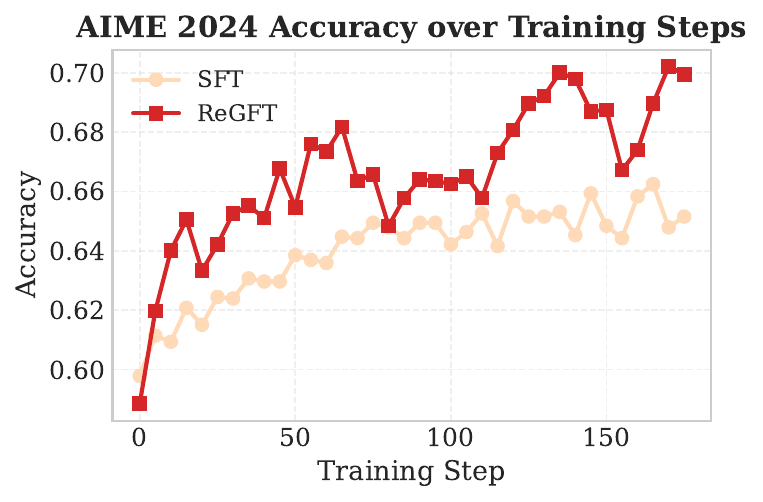}
        \caption{AIME 2024}
        \label{fig:sft_regft_aime24}
    \end{subfigure}
    \hfill
    \begin{subfigure}{0.32\textwidth}
        \centering
        \includegraphics[width=\linewidth]{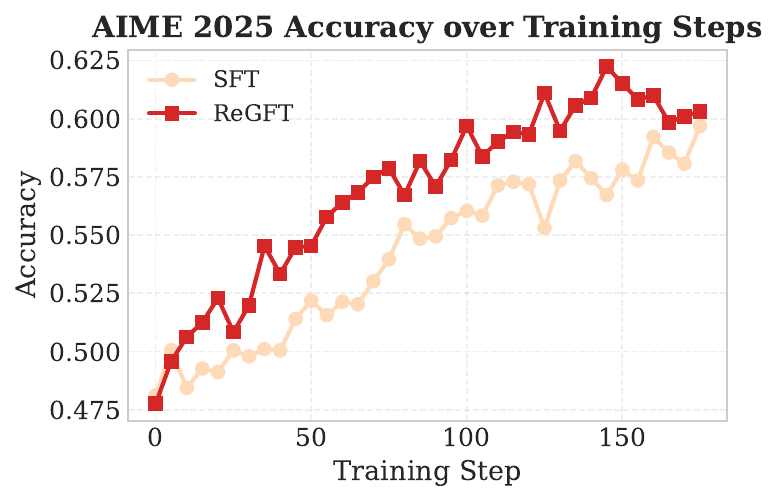}
        \caption{AIME 2025}
        \label{fig:sft_regft_aime25}
    \end{subfigure}
    \hfill
    \begin{subfigure}{0.32\textwidth}
        \centering
        \includegraphics[width=\linewidth]{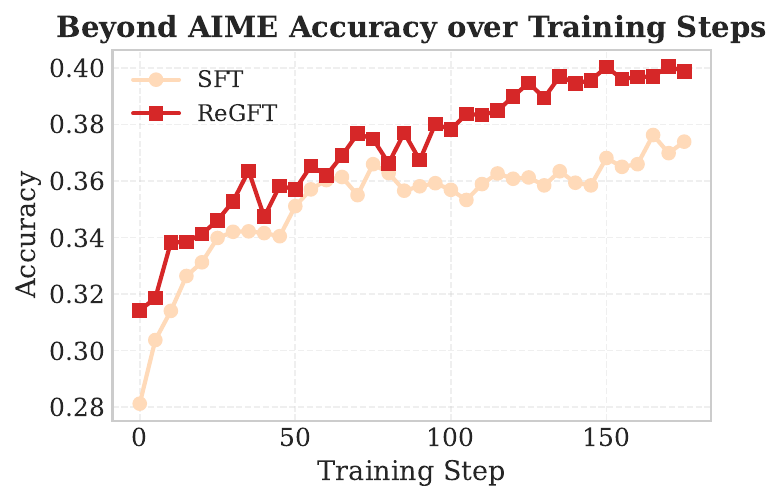}
        \caption{Beyond AIME}
        \label{fig:sft_regft_beyond}
    \end{subfigure}
    
    \vspace{-2mm}
    \caption{
        Accuracy comparison between direct SFT on raw human reference solutions and \textbf{ReGFT}.
        Models trained directly on reference solutions fail to achieve competitive RL performance,
        underscoring the importance of model-derived reasoning trajectories.
    }
    \label{fig:impact_model_comparison}
\end{figure*}

\subsection{Main Results}

\subsubsection{Reference-Guided Finetuning Enhances RL Training}

To evaluate whether reference-guided fine-tuning improves downstream reinforcement learning, we perform DAPO-based RL starting from both the original raw checkpoint and the ReGFT checkpoint. During training, we sample 64 responses per prompt to promote sufficient exploration of potentially correct reasoning trajectories; the effect of varying this sampling budget is analyzed separately in Section~\ref{sec:sample_scale}.

As shown in  Figure~\ref{fig:benchmark_comparison}, models initialized with ReGFT consistently outperform those initialized from the raw checkpoint across all three benchmarks throughout the entire RL process. Notably, the ReGFT-initialized models achieve higher accuracy at convergence and exhibit substantially faster improvement during the early and middle stages of training. This behavior indicates that ReGFT provides a stronger starting point for reinforcement learning, enabling more effective progress with fewer RL updates.

These results suggest that reference-guided fine-tuning alleviates the reward sparsity that commonly hinders reinforcement learning on hard reasoning tasks. By increasing the likelihood that sampled trajectories contain correct ones on hard problems, ReGFT supplies RL with more informative learning signals. As a result, RL converges more rapidly and reaches a higher performance ceiling, demonstrating that ReGFT improves not only optimization efficiency but also the attainable final performance under a strong RL algorithm.

\subsubsection{Impact of reference-Guided Demonstrations}

ReGFT differs from ReFT in that it incorporates reference-guided demonstrations in addition to self-generated correct trajectories, whereas ReFT relies exclusively on the model’s own successful rollouts. To isolate the contribution of human guidance, we perform DAPO-based reinforcement learning starting from both ReFT and ReGFT checkpoints.

As shown in Figure~\ref{fig:benchmark_comparison} and Figure~\ref{fig:impact_hint_comparison}, initializing RL from the ReFT checkpoint already leads to faster early-stage improvements compared to the raw model. This confirms that ReFT effectively densifies reward signals by increasing the probability of sampling correct trajectories, thereby improving optimization efficiency in the initial phase of RL. However, despite this advantage, ReFT consistently underperforms ReGFT at convergence across all benchmarks, and even show inferior performance compared with raw DAPO on BeyondAIME\citep{bytedance_seed_2025_beyondaime} benchmark.

The persistent performance gap indicates that ReFT primarily accelerates learning without fundamentally expanding the model’s reasoning capabilities. In contrast, ReGFT achieves higher asymptotic accuracy, suggesting that reference-guided demonstrations play a critical role beyond optimization speed. By introducing reasoning patterns that the model cannot reliably discover through self-exploration alone, human reference guidance enables ReGFT to extend the model’s competence to previously unsolvable problems. Consequently, ReGFT not only improves training efficiency but also raises the achievable performance limit after reinforcement learning.

\subsubsection{Necessity of Model-Derived Reasoning}

To further investigate how reference solutions should be incorporated, we compare ReGFT with a baseline that directly fine-tunes the model on raw human-written reference solutions, without requiring the model to generate its own reasoning. This ablation examines whether exposure to correct human reasoning alone is sufficient to improve downstream RL performance.

As shown in Figure~\ref{fig:impact_model_comparison}, directly fine-tuning on unprocessed human solutions leads to substantially weaker performance, both during supervised training and after reinforcement learning. This outcome indicates that simply providing correct reasoning traces is insufficient: the model struggles to internalize reasoning patterns that are not expressed in its own generative style or latent reasoning space.

In contrast, ReGFT prompts the model to produce its own reasoning trajectories under partial reference guidance, resulting in solutions that are both correct and distributionally aligned with the model’s inference process. These model-derived trajectories serve as an effective interface between human reference demonstrations and reinforcement learning. The results demonstrate that model-generated reasoning is a necessary component for successful knowledge transfer, and that ReGFT provides a principled mechanism for incorporating reference guidance without breaking this alignment.

\begin{figure*}[t]
\vspace{-2mm}
    \centering
    \begin{subfigure}{0.32\textwidth}
        \centering
        \includegraphics[width=\linewidth]{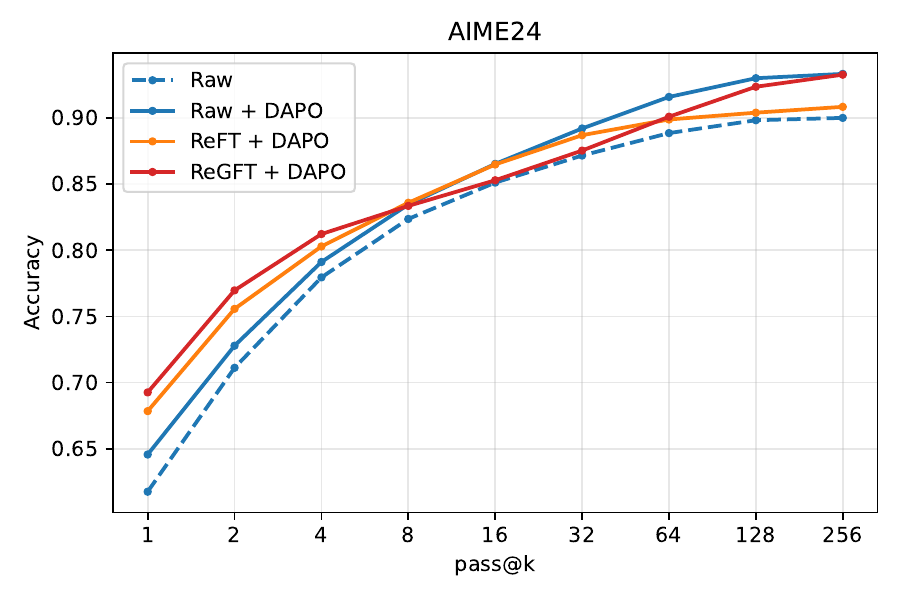}
        \vspace{-7mm}
        \caption{AIME 2024}
        \label{fig:testscaling_aime24}
    \end{subfigure}
    \hfill
    \begin{subfigure}{0.32\textwidth}
        \centering
        \includegraphics[width=\linewidth]{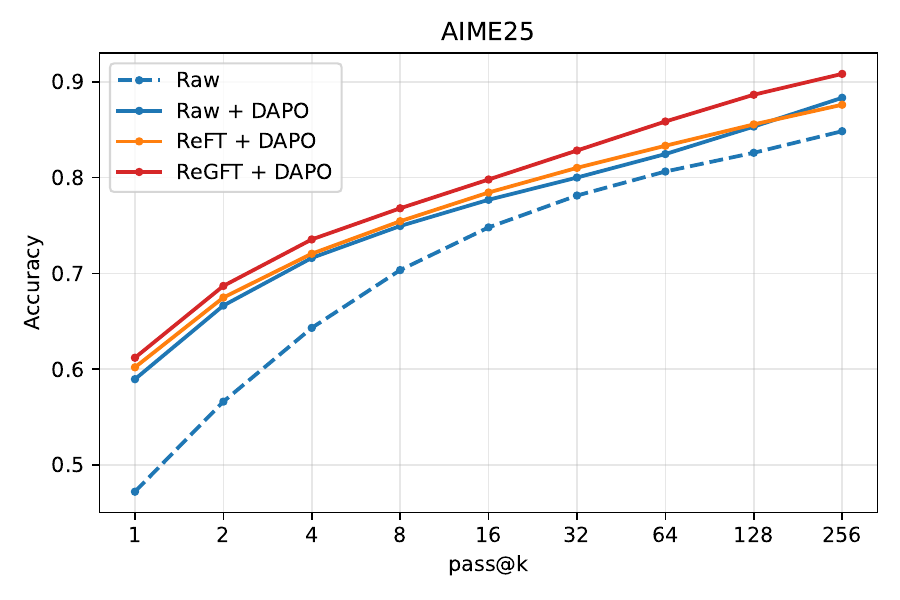}
        \vspace{-7mm}
        \caption{AIME 2025}
        \label{fig:testscaling_aime25}
    \end{subfigure}
    \hfill
    \begin{subfigure}{0.32\textwidth}
        \centering
        \includegraphics[width=\linewidth]{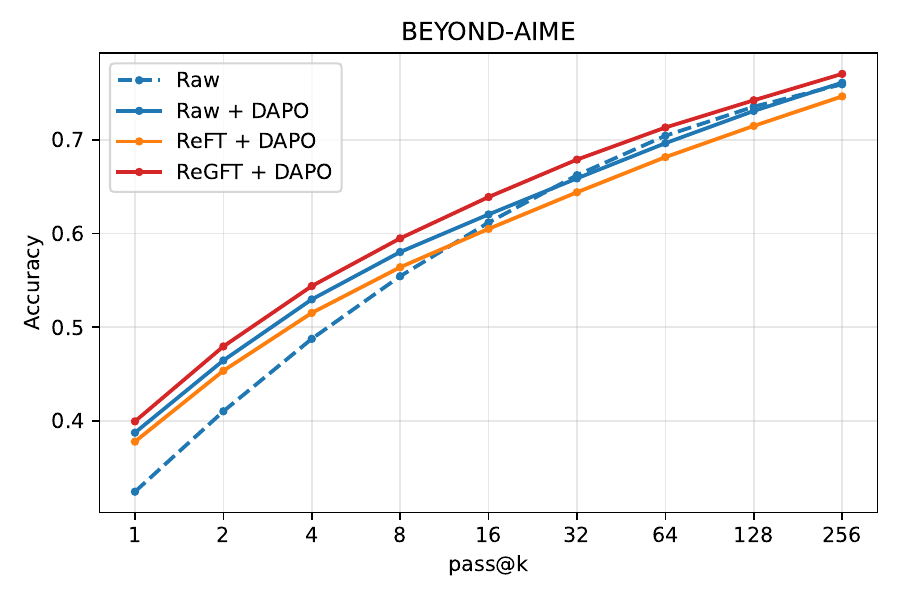}
        \vspace{-7mm}
        \caption{Beyond AIME}
        \label{fig:testscaling_beyondaime}
    \end{subfigure}
    \vspace{-2mm}
    \caption{
    Inference-time scaling performance (\texttt{pass@k}) of raw and RL-trained checkpoints across benchmarks.
    Solid lines denote DAPO-trained models, while dashed line indicate the raw pre-RL checkpoint.
    }
    \label{fig:test_time_scaling}
    \vspace{-2mm}
\end{figure*}

\subsubsection{Inference-Time Scaling Performance}

We evaluate inference-time scaling by measuring \texttt{pass@k} accuracy.
For each problem, we sample $N=1024$ independent generations and estimate \texttt{pass@k} as
\[
\mathrm{pass@k} \;=\; 1 - \frac{\binom{N - c}{k}}{\binom{N}{k}},
\]
where $c$ denotes the number of correct samples among the $N$ generations.

As shown in Figure~\ref{fig:test_time_scaling}, reinforcement learning consistently improves inference-time scaling behavior.
In contrast to prior observations that report early \texttt{pass@k} saturation \citep{yue2025doesreinforcementlearningreally}, we observe substantial gains across a wide range of $k$.
We attribute this to the difference of the base model, as well as we have a larger sampling budget during RL.

Across all benchmarks, \textbf{ReGFT + DAPO} exhibits the strongest and most consistent scaling performance.
It dominates ReFT- and raw-based baselines on AIME 2025 and Beyond-AIME, and outperforms them across both small and large $k$ regimes on AIME 2024.
Notably, ReGFT maintains a clear performance margin as $k$ increases, indicating improved coverage of the solution space rather than gains limited to a small number of samples.

In contrast, while ReFT improves \texttt{pass@1} over raw DAPO, these gains diminish as $k$ grows.
We hypothesize that this behavior arises because ReFT primarily reinforces the model’s existing successful trajectories, which can bias exploration and limit the diversity of solutions sampled at inference time.
By incorporating external reference guidance during pre-RL finetuning, ReGFT expands the model’s underlying competence on hard problems, leading to consistently higher \texttt{pass@k} under increased test-time compute.
These results highlight the effectiveness of ReGFT in enabling robust inference-time scaling beyond what can be achieved by self-reinforcement alone.

\begin{table*}[t]
\centering
\caption{Performance comparison of models fine-tuned under different settings. We report pass@64 accuracy on the training dataset and evaluation results on \textsc{AIME24}, \textsc{AIME25}, and \textsc{BeyondAIME}.}
\vspace{-0.5em}
\begin{tabular}{lccccc}
\toprule
\textbf{Model} & \textbf{Training Dataset (\%)} & \textbf{Train pass@64 (\%)} & \textbf{AIME24 (\%)} & \textbf{AIME25 (\%)} & \textbf{BeyondAIME (\%)} \\
\midrule
Raw        & 48.4 & 68.6 & 59.2 & 46.7 & 30.5 \\
ReFT        & 49.6 & 70.2 & 62.1 & 46.8 & 31.3 \\
ReGFT   & 50.1 & 72.5 & 60.0 & 47.8 & 31.3 \\
\bottomrule
\end{tabular}
\label{tab:training_results}
\end{table*}

\label{sec:sample_scale}

\begin{table}[t]
\centering
\vspace{-2mm}
\caption{Effect of sampling scale per problem under different training settings on AIME~2024, AIME~2025, and Beyond-AIME.}
\label{tab:main_results}
\vspace{-3mm}
\begin{tabular}{lccc}
\toprule
\textbf{Setting} & \textbf{AIME'24} & \textbf{AIME'25} & \textbf{Beyond-AIME} \\
\midrule
Raw model & 59.2 & 46.7 & 30.5 \\
\midrule
\multicolumn{4}{c}{\textbf{DAPO}} \\
\midrule
Response size 16 & 63.2 & 54.3 & 36.1 \\
Response size 64 & 67.1 & 60.5 & 39.8 \\
\midrule
\multicolumn{4}{c}{\textbf{ReFT + DAPO}} \\
\midrule
Response size 16 & 66.2 & 57.8 & 35.0 \\
Response size 64 & 68.3 & 60.8 & 38.3 \\
\midrule
\multicolumn{4}{c}{\textbf{ReGFT + DAPO}} \\
\midrule
Response size 16 & 67.0 & 56.7 & 37.4 \\
Response size 64 & \textbf{70.0} & \textbf{61.6} & \textbf{40.3} \\
\bottomrule
\end{tabular}
\end{table}

\subsection{Additional Experimental Evidence}

\subsubsection{Reference-Guided Generation Enables the Model to Solve Previously Unsolvable Problems}

We evaluate whether reference-guided generation can expand the set of problems for which the model is able to produce correct reasoning trajectories on the OmniMath dataset. We compare two sampling strategies: normal sampling and reference-guided sampling. For each problem, we sample 64 independent reasoning trajectories under each strategy. In the reference-guided condition, the model is provided with the first $80\%$ sentences of the human-written reference solution as contextual guidance, while the remaining $20\%$—which typically contains the final answer—is withheld. This design aims to avoid trivial answer copying, although in practice we found that the model almost always derives its own reasoning independently even when exposed to the full reference solution.

Across the 4{,}428 problems in OmniMath, standard sampling solves $68.58\%$ of the problems, whereas reference-guided sampling solves $70.82\%$. More importantly, reference-guided sampling enables the model to solve an additional $5.85\%$ of problems that are never solved under standard sampling, indicating that reference guidance can unlock reasoning trajectories that are otherwise inaccessible to the model. Conversely, $3.61\%$ of problems are solved only by standard sampling. In ReGFT, we train on correct solutions from both sampling strategies on these hard problems, thereby maximizing the probability of correct solution generation.

Despite access to human-written guidance, performance remains well below $100\%$. We identify two primary factors contributing to this gap. First, the model’s ability to interpret and internalize complex human reasoning is limited, particularly when reference solutions rely on advanced mathematical tools or domain-specific insights that fall outside the model’s training distribution. Second, the rule-based verifier used in OmniMath cannot reliably assess open-ended or proof-style solutions, leading to false negatives even when the model’s reasoning is substantively correct. These limitations highlight both the challenges of reference-guided generation and the constraints imposed by current automatic verification methods.

\subsubsection{Performance of the Fine-Tuned Models}

We evaluate the fine-tuned models obtained via ReFT and ReGFT on the OmniMath training set, as well as on challenging out-of-distribution benchmarks, including AIME 2024, AIME 2025, and Beyond-AIME. Table~\ref{tab:training_results} summarizes the results. Both fine-tuning strategies yield consistent performance improvements over the base model across the training set and all evaluation benchmarks, indicating that supervised fine-tuning on self-generated trajectories is effective for strengthening mathematical reasoning.

Comparing the two approaches, the ReGFT checkpoint achieves a higher pass rate on the OmniMath training set than the ReFT checkpoint. This improvement reflects the benefit of incorporating reference-guided trajectories, which expand the set of problems for which correct solutions are available during supervised fine-tuning. In contrast, performance on the external evaluation benchmarks remains largely comparable between ReFT and ReGFT, suggesting that the additional reference guidance does not lead to overfitting to the training distribution or degradation in generalization.

Taken together, these results indicate that ReGFT primarily improves the model’s ability to solve hard training problems by increasing the density and diversity of correct reasoning trajectories, while preserving robust generalization to unseen mathematical tasks. As a result, ReGFT provides a stronger initialization for subsequent reinforcement learning, where denser reward signals are critical for stable and effective optimization.

\subsubsection{Effect of Sampling Scale per Problem}

A common strategy for mitigating reward sparsity in reinforcement learning is to increase the number of sampled responses per problem, thereby raising the probability that at least one trajectory receives a positive reward \citep{hu2025brorlscalingreinforcementlearning}. In this ablation, we examine how sampling scale interacts with different model initializations by comparing response sizes of 16 and 64 under DAPO, ReFT+DAPO, and ReGFT+DAPO.

As shown in Table~\ref{tab:main_results}, increasing the response size from 16 to 64 consistently improves performance across all settings, confirming that larger sampling budgets help alleviate the zero-reward regime by providing denser learning signals. However, scaling alone is not sufficient: even with 64 responses per problem, models trained without reference-guided fine-tuning lag behind those initialized with ReGFT. In particular, ReGFT+DAPO achieves the strongest results across all benchmarks, indicating that improving the model’s initial competence and increasing exploration scale are complementary and jointly necessary for robust gains.

%% file: sections/conclusion.tex
\section{Conclusion}

This work addresses a central bottleneck in reinforcement learning for mathematical reasoning: reward sparsity caused by the absence of correct model-generated trajectories on hard problems. When a model cannot produce any rewarding samples, reinforcement learning stalls regardless of the algorithm or sampling budget.

We introduce Reference-Guided Fine-Tuning (ReGFT), which uses reference solutions to expand the model’s solvable problem set before reinforcement learning. Rather than training directly on reference chains of thought, ReGFT conditions the model on partial references and requires it to generate its own reasoning trajectories, producing solutions that are both correct and aligned with the model’s inference distribution. This increases the density and diversity of verifiable trajectories, enabling reinforcement learning to receive meaningful signals on problems that were previously unsolvable.

Across OmniMath, AIME 2024, AIME 2025, and Beyond-AIME, ReGFT improves supervised pass rates, accelerates DAPO training, and consistently yields higher final performance than both raw and ReFT-based baselines. Importantly, ReGFT also improves inference-time scaling: its advantage persists as k increases in pass@k evaluation, indicating more reliable discovery of correct solutions under additional test-time compute rather than gains confined to a small number of samples.

Overall, our results demonstrate that ReGFT mitigates reward sparsity by converting reference solutions into model-derived correct trajectories, yielding a stronger initialization that makes downstream RL more effective. The resulting models not only achieve higher accuracy after RL, but also exhibit stronger and more stable pass@k scaling.

%% file: sections/appendix.tex
\section{RL Training Implementation Details.}
Our reinforcement learning setup follows the DAPO\citep{yu2025dapo} framework, and we adopt its recommended hyperparameter configuration unless otherwise stated. We use the AdamW optimizer with a constant learning rate of $1\times10^{-6}$, together with a linear warm-up over the first 20 rollout steps. During rollout, we adjust the prompt batch size based on the number of sampled responses per prompt: when sampling 16 responses per prompt, we use a batch size of 512 prompts, and when sampling 64 responses per prompt, we use a batch size of 128 prompts. The mini-batch size is set to 2048, corresponding to 4 gradient updates per rollout step. For the Clip-Higher mechanism, we set the lower and upper clipping thresholds to $\varepsilon_{\text{low}}=0.2$ and $\varepsilon_{\text{high}}=0.28$, same as in the DAPO paper.

\section{Prompts.}

\begin{tcolorbox}[promptbox, title=Prompt for Solution Generation]
\ttfamily
Question: \{question\}

Please reason step by step, and put your final answer within \$boxed\{\}\$.
\end{tcolorbox}

\begin{tcolorbox}[promptbox, title=Prompt for Reference Guided Sampling]
\ttfamily
Question: \{question\}

Hint: \{partial reference solution\}

Given the partial reference solution known to be correct as hints, derive your solution to the question. You must solve it by yourself, and may follow the ideas in the hint. Please reason step by step, and put your final answer within \$boxed\{\}\$.
\end{tcolorbox}

\section{Problem Examples.}

We present representative training examples from OmniMath along with their corresponding truncated reference solutions used as guidance. The reference solutions are concise and omit detailed reasoning process, highlighting the need for reference-guided sampling to encourage the model to construct its own reasoning trajectories.

\begin{tcolorbox}[problembox, title=Problem in Omnimath 1]
\ttfamily
Suppose $a_i, b_i, c_i, i=1,2,\cdots ,n$, are $3n$ real numbers in the interval $\left [ 0,1 \right ].$ Define $$S=\left \{ \left ( i,j,k \right ) |\, a_i+b_j+c_k<1 \right \}, \; \; T=\left \{ \left ( i,j,k \right ) |\, a_i+b_j+c_k>2 \right \}.$$ Now we know that $\left | S \right |\ge 2018,\, \left | T \right |\ge 2018.$ Try to find the minimal possible value of $n$.
\end{tcolorbox}

\begin{tcolorbox}[problembox, title=Provided Partial Reference Solution 1]
\ttfamily
Suppose \( a_i, b_i, c_i \) for \( i = 1, 2, \ldots, n \) are \( 3n \) real numbers in the interval \([0, 1]\). Define the sets
\[
S = \{ (i, j, k) \mid a_i + b_j + c_k < 1 \}
\]
and
\[
T = \{ (i, j, k) \mid a_i + b_j + c_k > 2 \}.
\]
We are given that \( |S| \geq 2018 \) and \( |T| \geq 2018 \). We aim to find the minimal possible value of \( n \).

To establish a lower bound for \( n \), consider the projections of the sets \( S \) and \( T \) onto the coordinate planes. Note that \( S_{xy} \cap T_{xy} = \emptyset \), meaning that no pair \((a_i, b_j)\) can simultaneously satisfy \( a_i + b_j + c_k < 1 \) and \( a_i + b_j + c_k > 2 \) for any \( c_k \).

Thus, we have the inequalities:
\[
|S_{xy}| + |T_{xy}| \leq n^2, \quad |S_{yz}| + |T_{yz}| \leq n^2, \quad |S_{zx}| + |T_{zx}| \leq n^2.
\]

Applying the Projection Inequality and Hölder's Inequality, we obtain:
\begin{align*}
   &2 \cdot 2018^{2/3} \leq |S|^{2/3} + |T|^{2/3} \leq |S_{xy}|^{1/3} \cdot |S_{yz}|^{1/3} \cdot |S_{zx}|^{1/3} + |T_{xy}|^{1/3} \cdot |T_{yz}|^{1/3} \cdot |T_{zx}|^{1/3} \\ 
   &\leq (|S_{xy}| + |T_{xy}|)^{1/3} (|S_{yz}| + |T_{yz}|)^{1/3} (|S_{zx}| + |T_{zx}|)^{1/3} \leq n^2. 
\end{align*}

\end{tcolorbox}

\begin{tcolorbox}[problembox, title=Problem in OmniMath 2]
\ttfamily
For a given integer $n\ge 2$, let $a_0,a_1,\ldots ,a_n$ be integers satisfying $0=a_0<a_1<\ldots <a_n=2n-1$. Find the smallest possible number of elements in the set $\{ a_i+a_j \mid 0\le i \le j \le n \}$.
\end{tcolorbox}

\begin{tcolorbox}[problembox, title=Provided Partial Reference Solution 2]
\ttfamily
For a given integer \( n \ge 2 \), let \( a_0, a_1, \ldots, a_n \) be integers satisfying \( 0 = a_0 < a_1 < \ldots < a_n = 2n-1 \). We aim to find the smallest possible number of elements in the set \( \{ a_i + a_j \mid 0 \le i \le j \le n \} \).

First, we prove that the set \( \{ a_i + a_j \mid 1 \le i \le j \le n-1 \} \) takes all residues modulo \( 2n-1 \). Consider the \( 2n \) numbers:
\[ a_0 < a_1 < \cdots < a_{n-1} < a_n \]
and
\[ r - a_0 > r - a_1 > \cdots > r - a_{n-1} > r - a_n \]
for any integer \( 0 \le r \le 2n-2 \). By the Pigeonhole Principle, there must be two numbers that are congruent modulo \( 2n-1 \). Since \( a_i \not\equiv a_j \pmod{2n-1} \) for \( 1 \le i < j \le n-1 \), there exist \( 1 \le i, j \le n-1 \) such that \( a_i \equiv r - a_j \pmod{2n-1} \), meaning \( a_i + a_j \equiv r \pmod{2n-1} \).

Thus, the set \( \{ a_i + a_j \mid 1 \le i \le j \le n-1 \} \) takes all residues modulo \( 2n-1 \).

Returning to the original problem, we note that there are \( 2n+1 \) distinct numbers:
\[ a_0 + a_0 < a_0 + a_1 < \cdots < a_0 + a_{n-1} < a_n + a_0 < a_n + a_1 < \cdots < a_n + a_n, \]
which, modulo \( 2n-1 \), take only \( n \) different residues. Combining this with the fact that \( \{ a_i + a_j \mid 1 \le i \le j \le n-1 \} \) takes all residues modulo \( 2n-1 \), there are at least \( n-1 \) additional distinct numbers.

\end{tcolorbox}